\documentclass[a4paper,twoside]{article}

\usepackage{epsfig}
\usepackage{subcaption}
\usepackage{calc}
\usepackage{amssymb}
\usepackage{amstext}
\usepackage{amsmath}
\usepackage{amsthm}
\usepackage{multicol}
\usepackage{pslatex}
\usepackage{apalike}
\usepackage[bottom]{footmisc}
\usepackage{booktabs}  
\usepackage{url}  
\usepackage{graphicx}
\usepackage{tabularx}
\usepackage{amssymb}
\usepackage[ruled]{algorithm2e}
\usepackage{algpseudocode}
\usepackage{wrapfig}
\usepackage{caption}
\usepackage{amsmath}
\usepackage{amsthm}
\usepackage{textcomp}
\usepackage{xcolor}
\usepackage{tikz}
\usepackage{adjustbox}
\usetikzlibrary{arrows.meta}
\def\BibTeX{{\rm B\kern-.05em{\sc i\kern-.025em b}\kern-.08em
    T\kern-.1667em\lower.7ex\hbox{E}\kern-.125emX}}
    
\definecolor{painnasblue}{HTML}{0072B2}
\definecolor{painnasbluefill}{HTML}{DCEEF7}
\definecolor{painnasorange}{HTML}{D55E00}
\definecolor{painnasorangefill}{HTML}{FBE5D8}
\definecolor{painnasgreen}{HTML}{009E73}
\definecolor{painnasgreenfill}{HTML}{DDF3EB}
\definecolor{painnasyellow}{HTML}{E6B800}
\definecolor{painnasyellowfill}{HTML}{FFF3BF}
\definecolor{painnasgrey}{HTML}{6B7280}
\definecolor{painnaslightgrey}{HTML}{EEF0F2}
\definecolor{edacol}{HTML}{D55E00}
\definecolor{emgcol}{HTML}{009E73}
\definecolor{ecgcol}{HTML}{0072B2}
\definecolor{latentcol}{HTML}{6F5AA8}
\definecolor{neutralcol}{HTML}{59636E}
\definecolor{warningfill}{HTML}{FFF1E8}
\definecolor{goodfill}{HTML}{E8F5EF}
\definecolor{latentfill}{HTML}{EEEAF8}
\usepackage{SCITEPRESS}     

\begin{document}

\title{Efficient Architecture Search under Leave-One-Subject-Out Evaluation}

\author{\authorname{Heinke Hihn\sup{1}\orcidAuthor{0000-0002-3244-3661} and Friedhelm Schwenker \sup{2}\orcidAuthor{0000-0001-5118-0812}}
\affiliation{\sup{1}IT and Engineering Department, IU International University of Applied Sciences, Berlin, Germany}
\affiliation{\sup{2}Insitute for Neural Information Procressing, Ulm University, Ulm, Germany}
\email{heinke.hihn@iu.org, friedhelm.schwenker@uni-ulm.de}
}


\abstract{Deep neural architectures are widely used for signal processing in automated pain assessment systems. However, architecture design has remained largely a manual task despite the potential efficiency benefits of Neural Architecture Search (NAS). Embedding NAS in a Leave-One-Subject-Out (LOSO) evaluation is computationally demanding because a fully nested implementation requires $N$ independent architecture searches and, assuming approximately linear training cost, scales as $\mathcal{O}(N^2)$. We propose a block-based, leakage-controlled approach that shares NAS runs between subjects, reducing the number of searches from $N$ to $B$, where $B \ll N$, dubbed PainNAS. On the BioVid Heat Pain dataset, PainNAS yields comparable subject-level accuracy with substantially fewer parameters and FLOPs.}

\onecolumn \maketitle \normalsize \setcounter{footnote}{0} \vfill

\section{\uppercase{Introduction}}
\label{sec:introduction}
Automated pain assessment aims to build systems that classify the pain level a subject experiences~\cite{khan2025systematic} based on recordings of bodily reactions to a painful stimulus, such as EMG, EDA, and ECG data or video~\cite{khan2025systematic}. Recently, research has shifted from hand-crafted features to deep and hybrid approaches~\cite{khan2025systematic}. However, architecture design has remained a manual task despite the potential benefits of Neural Architecture Search (NAS)~\cite{ren2021comprehensive}. We argue that in part this is because of the search complexity introduced by the underlying evaluation strategy Leave-One-Subject-Out (LOSO) Cross Validation. LOSO is a commonly used subject-independent evaluation protocol because nested cross-validation separates model selection from performance
evaluation and thus reduces the optimistic bias that occurs when both are conducted using the same resampling results~\cite{cawley2010over}. As such, it is widely used in setting where the model evalaution must deal with subject-specific data, such as in Affective Computing~ \cite{hihn2016gestures,hihn2016inferring,thiam2019exploring,liu2024study,wang2026msfsnet}. In a leakage-controlled LOSO design for a dataset of $N$ subjects, a fully nested implementation would require $N$ independent architecture searches followed by an $N$-fold LOSO evaluation, thus growing in the order $\mathcal{O}(N^2)$. Therefore, an optimisation technique in this setting must support efficient LOSO evaluation.

\begin{figure}[t]
     \centering
\begingroup
\begin{adjustbox}{max width=\linewidth}

\begin{tikzpicture}[
  x=1cm,
  y=1cm,
  font=\scriptsize,
  >={Latex[length=1.35mm,width=1.05mm]},
  flow/.style={->, draw=painnasgrey, line width=0.45pt},
  subject/.style={
    draw=painnasblue,
    fill=painnasbluefill,
    rounded corners=0.5pt,
    minimum width=0.30cm,
    minimum height=0.30cm,
    inner sep=0pt,
    line width=0.45pt
  },
  minisubject/.style={
    subject,
    minimum width=0.20cm,
    minimum height=0.20cm
  },
  target/.style={
    draw=painnasorange,
    fill=painnasorangefill,
    rounded corners=0.5pt,
    minimum width=0.30cm,
    minimum height=0.30cm,
    inner sep=0pt,
    line width=0.60pt,
    text=painnasorange
  },
  minitarget/.style={
    target,
    minimum width=0.22cm,
    minimum height=0.22cm,
    font=\tiny
  },
  nas/.style={
    draw=painnasblue,
    fill=painnasbluefill,
    rounded corners=1.5pt,
    minimum width=1.12cm,
    minimum height=0.48cm,
    align=center,
    line width=0.55pt,
    font=\scriptsize\bfseries
  },
  mininas/.style={
    nas,
    minimum width=0.62cm,
    minimum height=0.32cm,
    inner sep=1pt,
    font=\tiny\bfseries
  },
  evaluation/.style={
    draw=painnasgrey,
    fill=painnaslightgrey,
    rounded corners=1.5pt,
    minimum width=1.38cm,
    minimum height=0.45cm,
    align=center,
    line width=0.45pt
  },
  statusbad/.style={
    draw=painnasorange,
    fill=painnasorangefill,
    rounded corners=2pt,
    inner xsep=3.5pt,
    inner ysep=1.8pt,
    text=painnasorange,
    font=\scriptsize\bfseries
  },
  statusgood/.style={
    draw=painnasgreen,
    fill=painnasgreenfill,
    rounded corners=2pt,
    inner xsep=3.5pt,
    inner ysep=1.8pt,
    text=painnasgreen,
    font=\scriptsize\bfseries
  }
]

\node[font=\scriptsize\bfseries, align=center]
  at (1.43,4.72) {(a) Global NAS};
\node[font=\tiny, text=painnasgrey]
  at (1.43,4.43) {search on all subjects};

\node[subject] (a1) at (0.43,3.91) {};
\node[subject] (a2) at (0.80,3.91) {};
\node[subject] (a3) at (1.17,3.91) {};
\node[font=\tiny, text=painnasgrey] at (1.47,3.91) {$\cdots$};
\node[subject] (an1) at (1.78,3.91) {};
\node[target]  (aj)  at (2.15,3.91) {$j$};
\node[font=\tiny, text=painnasgrey] at (2.49,3.91) {$\cdots$};
\node[font=\tiny, text=painnasgrey] at (1.43,4.17) {$\mathcal S$};

\node[nas] (globalnas) at (1.43,3.10) {NAS};
\draw[flow] (1.43,3.70) -- (globalnas.north);
\draw[->, draw=painnasorange, line width=0.60pt]
  (aj.south) .. controls (2.22,3.55) and (2.18,3.22) .. (globalnas.east);
\node[font=\tiny, text=painnasorange, anchor=west]
  at (2.10,3.49) {$j$ seen};

\node[evaluation] (globaleval) at (1.43,2.20) {LOSO evaluation};
\draw[flow] (globalnas) -- node[right, font=\tiny] {$a^\star$} (globaleval);

\node[font=\scriptsize\bfseries] at (1.43,1.52) {1 NAS search};
\node[statusbad] at (1.43,0.87) {optimistically biased};

\node[font=\scriptsize\bfseries, align=center]
  at (4.39,4.72) {(b) Nested LOSO NAS};
\node[font=\tiny, text=painnasgrey]
  at (4.39,4.43) {one search per target subject};

\node[font=\tiny, text=painnasgrey] at (3.38,4.08) {train};
\node[font=\tiny, text=painnasgrey] at (5.51,4.08) {test};

\node[minisubject] at (3.20,3.70) {};
\node[minisubject] at (3.45,3.70) {};
\node[minisubject] at (3.70,3.70) {};
\node[font=\tiny, text=painnasgrey] at (3.94,3.70) {$\cdots$};
\node[mininas] (nasone) at (4.56,3.70) {NAS};
\node[minitarget] (testone) at (5.51,3.70) {$1$};
\draw[flow] (4.06,3.70) -- (nasone.west);
\draw[flow] (nasone.east) -- (testone.west);

\node[minisubject] at (3.20,3.08) {};
\node[minisubject] at (3.45,3.08) {};
\node[minisubject] at (3.70,3.08) {};
\node[font=\tiny, text=painnasgrey] at (3.94,3.08) {$\cdots$};
\node[mininas] (nastwo) at (4.56,3.08) {NAS};
\node[minitarget] (testtwo) at (5.51,3.08) {$2$};
\draw[flow] (4.06,3.08) -- (nastwo.west);
\draw[flow] (nastwo.east) -- (testtwo.west);

\node[font=\normalsize, text=painnasgrey] at (4.39,2.58) {$\vdots$};

\node[minisubject] at (3.20,2.07) {};
\node[minisubject] at (3.45,2.07) {};
\node[minisubject] at (3.70,2.07) {};
\node[font=\tiny, text=painnasgrey] at (3.94,2.07) {$\cdots$};
\node[mininas] (nasn) at (4.56,2.07) {NAS};
\node[minitarget] (testn) at (5.51,2.07) {$N$};
\draw[flow] (4.06,2.07) -- (nasn.west);
\draw[flow] (nasn.east) -- (testn.west);

\node[font=\scriptsize\bfseries] at (4.39,1.43) {$N=87$ NAS searches};
\node[statusgood] at (4.39,0.87) {leakage-free};

\node[font=\scriptsize\bfseries, align=center]
  at (7.33,4.72) {(c) PainNAS};
\node[font=\tiny, text=painnasgrey]
  at (7.33,4.43) {$B=5$ outer blocks};

\node[subject, minimum width=0.40cm, font=\tiny] (g1) at (6.26,3.98) {$\mathcal G_1$};
\node[subject, minimum width=0.40cm, font=\tiny] (g2) at (6.79,3.98) {$\mathcal G_2$};
\node[target,  minimum width=0.40cm, font=\tiny] (gb) at (7.32,3.98) {$\mathcal G_b$};
\node[subject, minimum width=0.40cm, font=\tiny] (g4) at (7.85,3.98) {$\mathcal G_4$};
\node[subject, minimum width=0.40cm, font=\tiny] (g5) at (8.38,3.98) {$\mathcal G_5$};
\node[font=\tiny, text=painnasorange]
  at (7.32,3.64) {excluded};

\node[font=\tiny, text=painnasgrey]
  at (7.33,3.35) {$\mathcal T_b=\mathcal S\setminus\mathcal G_b$};

\draw[draw=painnasgrey!75, fill=painnaslightgrey!55,
      rounded corners=2pt, line width=0.45pt]
  (6.22,2.05) rectangle (8.44,3.17);
\node[font=\tiny\bfseries, text=painnasgrey]
  at (7.33,3.02) {inner $K=3$ folds};

\node[font=\tiny, text=painnasgrey] at (6.70,2.79) {$k=1$};
\node[font=\tiny, text=painnasgrey] at (7.33,2.79) {$k=2$};
\node[font=\tiny, text=painnasgrey] at (7.96,2.79) {$k=3$};

\node[minisubject, fill=painnasyellowfill, draw=painnasyellow] at (6.70,2.58) {};
\node[minisubject] at (6.70,2.39) {};
\node[minisubject] at (6.70,2.20) {};
\node[minisubject] at (7.33,2.58) {};
\node[minisubject, fill=painnasyellowfill, draw=painnasyellow] at (7.33,2.39) {};
\node[minisubject] at (7.33,2.20) {};
\node[minisubject] at (7.96,2.58) {};
\node[minisubject] at (7.96,2.39) {};
\node[minisubject, fill=painnasyellowfill, draw=painnasyellow] at (7.96,2.20) {};

\node[nas, minimum width=1.22cm] (blocknas) at (7.33,1.58) {NAS$_b$};
\draw[flow] (7.33,2.05) -- (blocknas.north);
\node[font=\tiny, text=painnasgrey, anchor=west]
  at (7.99,1.58) {$\times 5$};

\node[font=\scriptsize\bfseries] at (7.33,1.02) {$B=5$ NAS searches};
\node[statusgood] at (7.33,0.54) {leakage-free};
\end{tikzpicture}
\end{adjustbox}
\endgroup
\caption{\textbf{(a)} A
    single global search uses all subjects. \textbf{(b)} Fully nested LOSO NAS excludes
    every target subject from its corresponding search. \textbf{(c)} The proposed approach excludes blocks-wise, evaluates candidates using subject-disjoint inner folds, and shares architectures.}
    \label{fig:nas-loso-comparison}
  \end{figure}
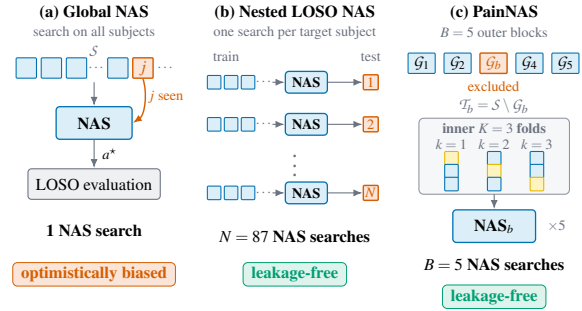

Deep architectures in automated pain assessment can be categorised in early fusion (EF) and late fusion (LF). Early fusion approaches are comparatively under-represented (e.g., the review by ~\cite{khan2025systematic} identifies a single early-fusion method), and several published comparisons report stronger late- or intermediate-fusion results~\cite{thiam2019exploring}. Consequently, studies have often used modality-specific encoders with adaptive weighting, gating, or attention~\cite{khan2025systematic}. One possible reason for this is that raw concatenation requires heterogeneous physiological signals to share a common representation which is rarely the case, whereas late-fusion models use separate processing paths before combining representations or decisions into a common learned feature space. However, it remains unclear whether their advantage arises from the fusion principle itself or from their more specialised and extensively engineered modality-specific encoders. A suitably selected shared architecture may compensate for the heterogeneous temporal characteristics of the physiological modalities while retaining the efficiency advantages of early fusion.

Based on this, we formulate the following research questions: 
\begin{enumerate}
    \item[\textit{RQ1:}]  To what extent can leakage-free neural architecture search improve subject-independent pain recognition compared to a manually designed architecture? 
    \item[\textit{RQ2:}] What trade-off does PainNAS provide between search cost and LOSO generalisation performance?, and 
    \item[\textit{RQ3:}] Which architectural characteristics are consistently selected across subject blocks and classification settings?
\end{enumerate}

To answer these questions, we introduce PainNAS: a cross-fitted block NAS procedure that excludes every evaluated subject from the architecture search used for that subject, while reducing the number of NAS runs, thus enabling an efficient and leakage-free NAS within a LOSO-CV.

\section{\uppercase{Related Work}}
\label{sec:related_work}
We divide previous work into two categories: \textit{(i)} NAS on physiological signals, and \textit{(ii)} efficient NAS methods. To the best of our knowledge, there is no prior work in the literature introducing a design for subject-wise outer evaluation which finds a leakage-free architecture.

\subsection{NAS for Physiological Signal Processing}
\label{sec:nas_physio}
NAS has been applied to physiological signals
and multimodal fusion~\cite{wu2023autoeer,ghebriout2024harmonic},
but these methods do not address the repeated architecture searches required by subject-wise outer evaluation. In contrast, our approach reduces the number of NAS runs by allowing subjects in the same outer block to share an architecture search from which the entire block is excluded. 

\subsection{Efficient NAS}
\label{sec:efficient_nas}
NAS can be efficiently implemented in several ways, the most common include sharing weights among candidate architectures \cite{pham2018efficient}, by using zero-cost proxies
to efficiently score architectures~\cite{abdelfattah2021zero}, and by guiding the search to promising avenues using, e.g., generative models~\cite{lukasik2022learning}. Such approaches would reduce the complexity of the LOSO-looped NAS but cannot find a leakage-free architecture by design.
\section{\uppercase{Method}}
\label{sec:method}

\begin{algorithm}[t]
\caption{PainNAS: cross-fitted block NAS with final LOSO evaluation}
\label{alg:painnas-cross-fitted-loso}
\small
\KwIn{Subjects $\mathcal S$, number of outer blocks $B$, number of inner folds $K$}
\KwOut{Mean LOSO accuracy}

Split $\mathcal S$ into disjoint outer blocks
$\mathcal G_1,\ldots,\mathcal G_B$\;

\ForEach{outer block $\mathcal G_b$}{
    Exclude $\mathcal G_b$ and split
    $\mathcal S\setminus\mathcal G_b$ into inner folds
    $F_1,\ldots,F_K$\;

    \ForEach{candidate architecture $a\in\mathcal A$}{
        \ForEach{inner fold $F_i$}{
            Fit $a$ using $\bigcup_{\ell\neq i}F_\ell$\;
            Evaluate $a$ on $F_i$\;
        }
        Compute the objective $J_b(a)$ from the inner-fold results\;
    }

    $a_b^\star \gets \arg\max_{a\in\mathcal A} J_b(a)$\;

    \ForEach{LOSO target subject $j\in\mathcal G_b$}{
        Fit $a_b^\star$ on $\mathcal S\setminus\{j\}$\;
        Evaluate once on the held-out subject $j$\;
    }
}
\Return mean LOSO accuracy\;
\end{algorithm}
This section introduces our method PainNAS as a LOSO evaluation and its integration with a computationally feasible, leakage-free NAS protocol.

\begin{table}[t]
    \centering
    \caption{%
        Results on the BioVid dataset in the binary and MC setting. Standard deviations are shown in parentheses where available. Bold indicates the best reported value.
    }
    \label{tab:pain_recognition_overview}
    \setlength{\tabcolsep}{3pt}
    \begin{adjustbox}{max width=\linewidth}
    \begin{tabular}{lll}
        \toprule
        \multicolumn{3}{c}{\textsc{Late Fusion}}\\
        \midrule
        \textbf{Method} & \textbf{Binary} & \textbf{Multi-Class} \\
        \midrule
        \cite{CrossModTransformer2025}
        & \textbf{87.52\% ($\pm 11.0$)} & N/A \\

        \cite{li2024automatic} 
        & 86.21\%  
        & 38.03\% \\
        
        \cite{Lu2023PainAttnNet} 
        & 85.56\%  
        & 34.46\% \\

        \cite{thiam2025dealing}
        & 85.32\% ($\pm 13.7$) & N/A \\

        \cite{jiang2024personalized}
        & 84.58\% ($\pm 13.3$)
        & \textbf{39.24\%} ($\pm 8.6$) \\

        \cite{thiam2019exploring}
        & 84.40\% ($\pm 14.4$)
        & 36.54\% ($\pm 8.5$) \\

        \cite{thiam2021multi}
        & 84.20\% ($\pm 13.7$) & 35.50\% ($\pm 7.9$) \\

        \cite{Steur2025Supervised}
        & 84.22\% ($\pm 13.2$) & N/A \\
        \midrule
        PainNAS
        & 83.39\% ($\pm 14.4$)
        & 35.82\% ($\pm 9.6$) \\
        Global NAS
        & 84.00\% ($\pm 15.8$)
        & 36.20\% ($\pm 9.7$) \\
        $\beta = 0$ baseline
        & 83.41\% ($\pm 14.4$)
        & 35.51\% ($\pm 9.4$) \\
        \midrule
        \multicolumn{3}{c}{\textsc{Early Fusion}}\\
        \midrule
        \textbf{Method} & \textbf{Binary} & \textbf{Multi-Class} \\
        \midrule
        \cite{thiam2019exploring}
        & 82.79\% ($\pm 15.2$) & N/A \\

        \cite{werner2016automatic}
        & 72.4\%  & 30.8\% \\
        \cite{werner2014automatic}
        & 77.8\%  & N/A \\
        \cite{kachele2016methods}
           & 82.7\% & 33.0\% \\
        \midrule
        PainNAS
        & \textbf{83.71\%} ($\pm 14.5$)
        & \textbf{35.84\%} ($\pm 9.6$) \\
        Global NAS
        & 82.72\% ($\pm 15.6$) 
        & 35.64\% ($\pm 9.3$) \\
        NAS per LOSO-fold
        & 82.35\% ($\pm 15.5$)
        & 34.39\% ($\pm 9.0$) \\
        $\beta = 0$ baseline
        & 82.29\% ($\pm 14.4$)
        & 35.26\% ($\pm 9.1$) \\
        \bottomrule
    \end{tabular}
    \end{adjustbox}
\end{table}

Formally, the accuracy of Leave-One-Subject-Out cross-validation is
\begin{equation}
  \operatorname{Acc}_{\mathrm{LOSO}}  =\frac{1}{N}\sum_{j=1}^{N}\operatorname{Acc}\!\left(D_{-j},D_j\right),
\end{equation}
where \(\operatorname{Acc}\!\left(D_{-j},D_j\right)\) denotes the classification accuracy of a model trained in dataset \(D_{-j}\), which excludes subject \(j\), and evaluated on the data of that subject \(D_j\). Given \(N\) subjects, NAS and final model fitting are performed independently for each of the \(N\) folds using the remaining \(N-1\) subjects. To avoid data leakage, no samples of the held-out subject are used during optimisation. Thus, the total runtime is proportional to 

\begin{align}
\begin{split}
 T_{\mathrm{LOSO\text{-}NAS}} &= N[T_{\mathrm{NAS}}(N-1) \\ & +T_{\mathrm{fit}}(N-1) + T_{\mathrm{test}}(1)] \\ & \in \mathcal{O}(N^2),
\end{split}
\end{align}

assuming \(T_{\mathrm{NAS}}(N),T_{\mathrm{fit}}(N)\in\mathcal{O}(N)\). To reduce the number of NAS runs, we divide the subjects $\mathcal{S}$ into disjoint \textit{outer} blocks $\mathcal{G}_1,\ldots,\mathcal{G}_B$. For each outer block $b$, NAS uses only the training set $\mathcal{T}_b=\mathcal{S}\setminus\mathcal{G}_b$. To evaluate a proxy for LOSO generalisation, we further split $\mathcal{T}_b$ into $K$ subject-disjoint inner folds: For each candidate architecture $a$ and inner fold $k$, a model is initialised independently, fitted on the other \(K-1\) folds, and validated on fold $k$. While we can reduce the number of NAS runs to $B$, the overall complexity still grows approximately quadratic in the number of subjects $N$ as we still need to do $N$ model fits and $N$ evaluations, 
\begin{align}
\begin{split}
    T_{\mathrm{PainNAS}} &= B\,T_{\mathrm{NAS}}\!\left(N-\frac{N}{B}\right) \\ & + N\left[\,T_{\mathrm{fit}}(N-1)+T_{\mathrm{test}}(1)\right]\in \mathcal{O}(N^2).
\end{split}
\end{align}
The selection objective used the rank architectures during NAS is given as
\begin{align}
\bar A_b(a) &= \frac{1}{N_b}\sum_{i\in\mathcal{T}_b}A_i(a),\\
\widehat{\sigma}_b(a) &=
\sqrt{\frac{1}{N_b-1}\sum_{i\in\mathcal{T}_b}
\left(A_i(a)-\bar A_b(a)\right)^2},\\
J_b(a) &=\bar A_b(a)-\beta
\frac{\widehat{\sigma}_b(a)}{\sqrt{N_b}},
\label{eq:objective}
\end{align}
where $\bar A_b(a)$ is the mean subject-level validation accuracy of architecture \(a\) over $\mathcal{T}_b$, $N_b=|\mathcal{T}_b|$ is the number of subjects in $\mathcal{T}_b$, $\widehat\sigma_b(a)$ is the sample standard deviation of the subject-level accuracies over $\mathcal{T}_b$, and $J_b(a)$ defines a selection score. It rewards mean subject-level accuracy and penalises estimates that exhibit high inter-subject variance controlled by a factor $\beta$. We used $\beta = 1.0$. Algorithm~\ref{alg:painnas-cross-fitted-loso} gives a high-level overview.

After the inner folds, we retain the checkpoint with the highest validation score of the architecture found by optimising Eq.~\ref{eq:objective}. The model is then trained for additional \(M)\) epochs using the samples of all 86 non-target subjects, including the former inner-validation subjects and the other members of $\mathcal{G}_b$. We set \(M\) as the median number of epochs used in the NAS search before early-stopping. Finally, it is evaluated once on the held-out data $D_j$. The final reported metric is the mean accuracy across all target subjects.

\section{\uppercase{Results}}
\label{sec:results}
\begin{table}[t]
    \centering
    \caption{Results of blocked NAS over early fusion integrated with LOSO evaluation for different numbers of outer blocks and inner folds.}
    \begin{adjustbox}{max width=\linewidth}
    \begin{tabular}{lllll}
    \toprule
    \multicolumn{5}{c}{\textsc{Outer Blocks with K = 3}} \\
    \midrule
    & \multicolumn{2}{c}{\textbf{Binary}} & \multicolumn{2}{c}{\textbf{Multi-Class}} \\
    \midrule
    $B$ & \textbf{Accuracy} & \textbf{F1} & \textbf{Accuracy} & \textbf{F1} \\
    \midrule
    0 & 82.7\% ($\pm 15.6$) & 81.3\% ($\pm 14.0$) & 35.6\% ($\pm 9.3$) & 31.0\% ($\pm 9.3$) \\
    5 & 83.3\% ($\pm 14.4$) & 82.3\% ($\pm 16.4$) & 35.0\% ($\pm 9.5$) & 31.2\% ($\pm 9.8$)  \\
    10 & 83.8\% ($\pm 14.5$) & 82.6\% ($\pm 16.8$) & 35.8\% ($\pm 9.6$) & 31.2\% ($\pm 10.0$) \\
    20 & 83.7\% ($\pm 14.5$) & 82.6\% ($\pm 16.7$) & 35.6\% ($\pm 9.1$) & 31.0\% ($\pm 8.8$) \\
    87 & 82.3\% ($\pm 15.5$) & 81.2\% ($\pm 17.5$) & 34.4\% ($\pm 9.4$) & 30.4\% ($\pm 9.1$) \\
    \midrule
    \multicolumn{5}{c}{\textsc{Inner Folds with $B$ = 5}} \\
    \midrule
    & \multicolumn{2}{c}{\textbf{Binary}} & \multicolumn{2}{c}{\textbf{Multi-Class}} \\
    \midrule
    $K$ & \textbf{Accuracy} & \textbf{F1} & \textbf{Accuracy} & \textbf{F1} \\
    \midrule
    3 & 83.3\% ($\pm 14.4$) & 82.3\% ($\pm 16.4$) & 35.0\% ($\pm 9.5$) & 31.2\% ($\pm 9.8$)  \\
    5 & 82.3\% ($\pm 15.4$) & 82.6\% ($\pm 17.3$) & 35.6\% ($\pm 8.8$) & 30.8\% ($\pm 9.0$) \\
    10 & 82.7\% ($\pm 15.1$) & 81.4\% ($\pm 12.5$) & 34.5\% ($\pm 9.3$) & 30.0\% ($\pm 9.4$) \\
    \bottomrule
    \end{tabular}
    \end{adjustbox}
    \label{tab:block_count}
\end{table}
\begin{figure*}[ht]
    \centering
    \includegraphics[width=\linewidth, trim={0.25cm 0.25cm 0.25cm 0.25cm}, clip]{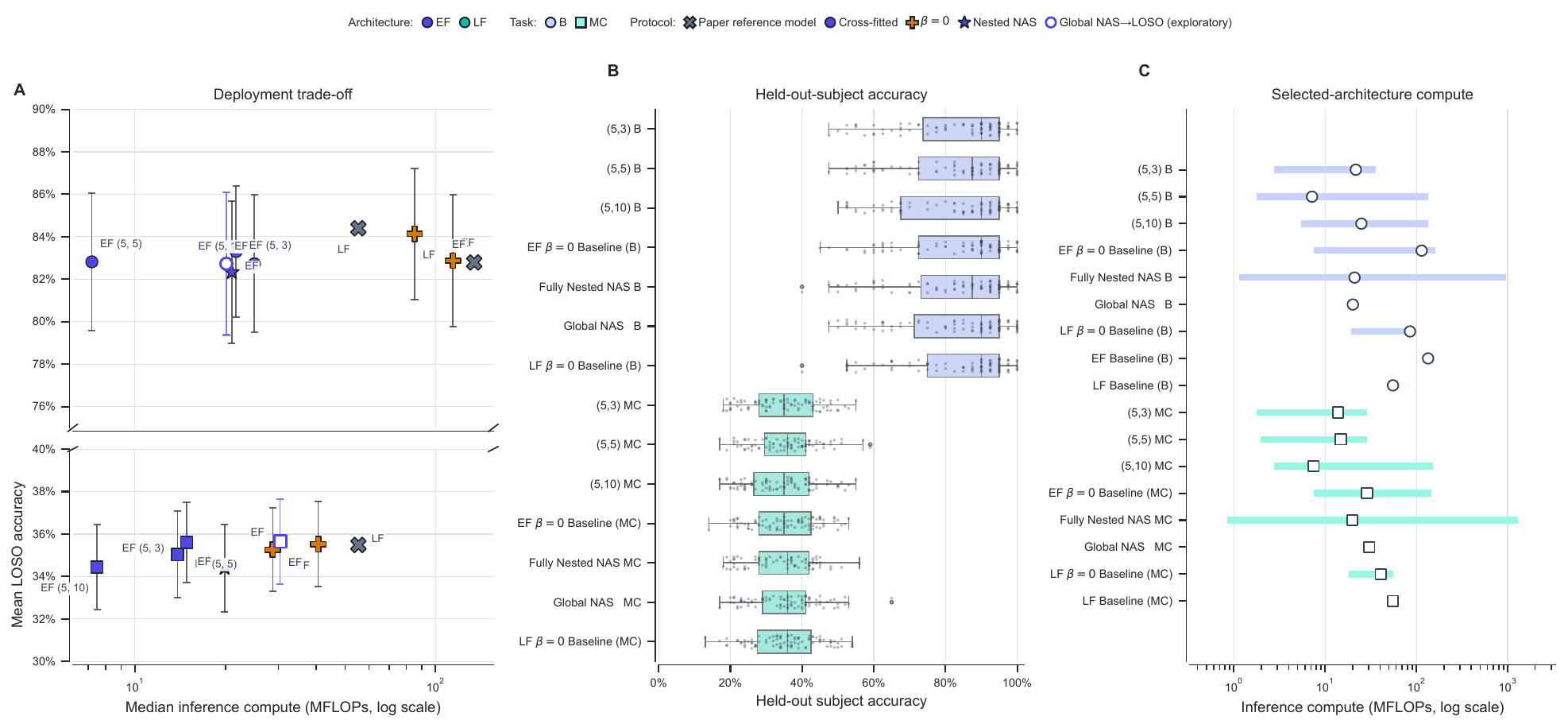}
    \caption{(A) Mean LOSO accuracy versus median inference compute for PainNAS, global-
  NAS--then--LOSO runs, and fixed early-/late-fusion baselines (B) Distribution of held-out-subject accuracy for each setting. (C) Range and median of the
  compute of architectures selected within each setting. $(i,j)$ indicates PainNAS with $i$ outer blocks and $j$ inner folds.}
    \label{fig:painnas_tradeoff}
\end{figure*}

In the following we present and discuss the results of the experiments we have designed to answer the research questions posed in Section~\ref{sec:introduction}.

\subsection{Classification Results}
To answer RQ1, we evaluated the BioVid heat pain dataset~\cite{Walter2013BioVid} and used the architecture proposed by~\cite{thiam2019exploring} as a baseline. To this end, Table~\ref{tab:pain_recognition_overview} compares with previous results in BioVid. Within our early-fusion protocol, PainNAS has a $+1$ pp higher mean binary accuracy than the manually designed baseline, while the selected architectures contain fewer parameters. In the multi-class (MC) setting, our early fusion architecture outperforms previous methods by a margin of $2.8$ pp. In the LF setting, PainNAS yields results comparable to previous methods while drastically reducing the number of parameters (see Section~\ref{sec:tradeoff}).

\subsection{Computational Resources and Trade-Offs}
\label{sec:tradeoff}
RQ2 investigates how PainNAS compares with a fully informed (but biased) NAS followed by LOSO evaluation. 

Investigating the effect of a larger block count is not straightforward. With five outer blocks and three inner folds, 17--18 subjects are excluded from search, with 10 and 20 blocks this becomes 8--9 and 4--5, respectively. As the number of blocks increases, architecture selection uses more subjects and the transferred warm-start checkpoint has seen more subjects. Table~\ref{tab:block_count} reports the results for different block counts under binary and MC settings, indicating PainNAS yields stable results and varying \(B\) and \(K\) settings.

Figure~\ref{fig:painnas_tradeoff} summarises the accuracy--compute trade-off across PainNAS settings. Binary classification consistently achieved high LOSO accuracy of approximately 80--85\%, and MC performance was roughly 30--40\%. The global-NAS--then--LOSO results should be interpreted as exploratory because their architecture search used cohort-wide information, but nevertheless, they show a similar task-dependent pattern.

Using an analytical estimate of two FLOPs per convolutional or dense operation, the manually designed binary early-fusion baseline requires \(134.42\,\mathrm{MFLOPs}\) and achieves \(82.79\%\) accuracy. Across the five PainNAS configurations, the median selected-architecture cost decreases to \(7.19\)--\(24.96\, \mathrm{MFLOPs}\), corresponding to \(5.4\times\)--\(18.7\times\) fewer FLOPs, while subject-averaged accuracy remains between \(82.73\%\) and \(83.71\%\). The \(B=20,K=3\) configuration has the highest observed point estimate, \(83.71\%\) accuracy at \(8.80\,\mathrm{MFLOPs}\), a \(15.3\times\) reduction relative to the baseline. MC accuracy is likewise stable at \(35.03\%\)--\(35.78\%\) over median costs of \(13.85\)--\(20.41\,\mathrm{MFLOPs}\).

\begin{figure}[t]
    \centering
    \includegraphics[width=\linewidth,trim={0cm 0.5cm 14cm 0.55cm}, clip]{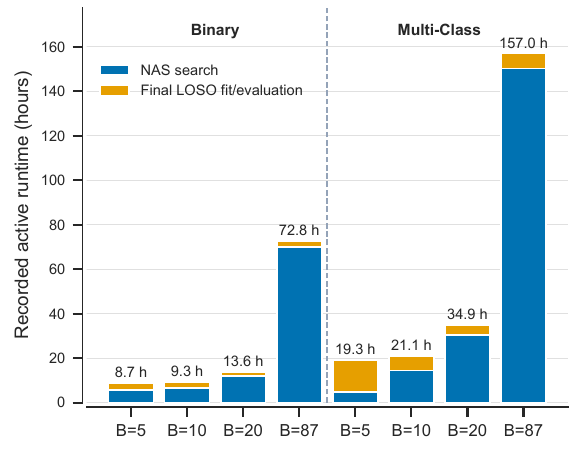}
    \caption{Runtime on a 24 GB L4 GPU recorded for different outer block sizes $B$ demonstrating the benefits of PainNAS. $B=87$ represents a fully LOSO-nested NAS.}
    \label{fig:nas_runtimes}
\end{figure}

The active run-time on a 24 GB GPU for the fully nested binary NAS was 72.81h, while PainNAS with five outer blocks and three inner folds took 8.69h under the exact same settings. In the fully nested MC run, it tool 157h compared to 19.3 using five blocks -- see Figure~\ref{fig:nas_runtimes} for more details. The fully nested run is roughly 8 times slower in both task setting but the results are comparable: 83.71\% and 35.84\%  for PainNAS and 82.34\% and 35.64\% accuracy for the fully nested NAS. No statistically significant difference was observed.

\begin{figure}[t]
    \centering
    \includegraphics[width=0.925\linewidth, trim={0cm 0cm 18.5cm 0cm}, clip]{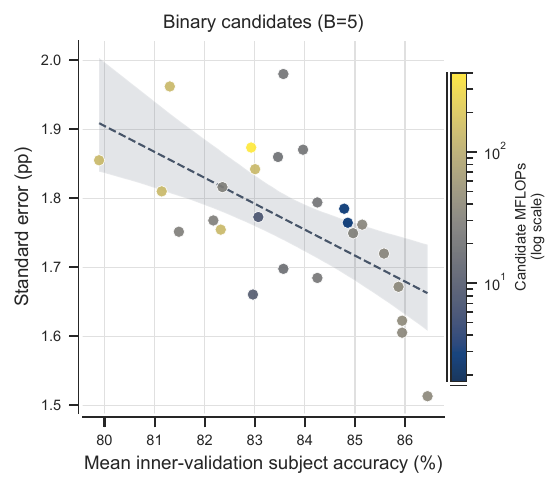}
    \caption{The relationship between the standard error used in the selection objective Eq.~\ref{eq:objective}, the MFLOPs, and the mean accuracy of the selected architectures indicates that minimising $\widehat{\sigma}_b(a)$ encourages finding efficient architectures.}
    \label{fig:uncertainty_objective}
\end{figure}

We also investigated the effect of the objective Eq.~\ref{eq:objective}. Compared with early fusion baselines, the full variance-penalised runs ($\beta=1.0$) selected substantially more efficient architectures. For binary classification, median inference cost decreased from $114.21$ to $21.70$~MFLOPs (an $81.0\%$ reduction). For MC classification, the median decreased from $28.74$ to $13.85$~MFLOPs (a $51.8\%$ reduction). This pattern did not extend to late fusion (LF): the binary median increased from $85.17$ to $123.97$MFLOPs (a $45.5\%$ increase), and the MC median increased from $40.71$ to $93.13$MFLOPs (a $128.7\%$ increase). Thus, in these runs, the variance-penalised objective’s compute benefit was specific to early fusion rather than a fusion-architecture-independent effect -- see Table~\ref{tab:pain_recognition_overview} and Figure~\ref{fig:painnas_tradeoff} for results and Figure~\ref{fig:uncertainty_objective} for the effect of the objective on architecture selection.

Thus, our results motivate PainNAS as a resource-aware approach to multichannel biomedical signal processing.

\subsection{Architectural Characteristics}
To answer RQ3, we examined the architectures selected by NAS across the five outer-block winners. Among the ten early fusion winners, separable convolutions were selected in $9/10$ cases, group normalisation in $7/10$, average pooling in $7/10$, and pooling size four in $9/10$, while batch-dependent normalisation was selected only once. MC selected global-average heads more often than binary models ($3/5$ versus $2/5$), but selected four-block architectures in only $3/5$ cases (the remaining used three blocks), whereas all binary winners used four or five blocks. Temporal kernels of $15$ were selected equally often in both settings ($2/5$), and width multiplier $2.0$ was selected less often for MC than binary models ($2/5$ vs. $3/5$).

In LF, the EDA branch used a flatten head in all ten selected architectures. Binary models selected global-average heads more frequently for ECG ($4/5$ vs. $2/5$ for MC) and EMG ($3/5$ vs. $2/5$). Conversely, MC late-fusion branches more often had six or seven blocks ($13/15$ branches, compared with $10/15$ for binary) and used flatten heads more frequently overall ($11/15$ vs. $8/15$). The LF reference baseline~\cite{thiam2019multi} used seven-block flatten branches for all modalities.

The selections suggest that robust pain recognition may benefit from architectures that reduce temporal resolution early and avoid batch-dependent normalisation, which can be unstable when training data are limited or distributed across subjects. 
The MC preference for wider networks, longer kernels, and global-average heads suggests that distinguishing several pain levels requires broader temporal context but benefits from a more compact final representation. 

\subsection{Experimental Setup}
\label{sec:setup}
We used Optuna~\cite{akiba2019optuna} with a TPE sampler for architecture search varying the number of convolutional blocks (3--5), repetitions per block (1--2), width multiplier ($0.5$, $1$, or $2$), kernel size (7, 11, or 15), convolution type, normalisation and pooling types, classifier depth and width, and learning rate ($10^{-5}$--$10^{-3}$, log scale). In the LF setting, the architectures of the EMG and ECG branches were mirrored to reduce the search space. Models were trained for at most 30 epochs using a batch size of 128 and early stopping. Successive-halving pruning removed weak trials after intermediate folds. The final objective was given by Eq.~\eqref{eq:objective}. The implementation is available at \url{https://github.com/xxx/xxx}.

\section{\uppercase{Discussion}}
\label{sec:discussion}
We have proposed a method to implement a computationally feasible NAS in the LOSO setting that improved the baseline and substantially reduced the number of parameters, increasing model efficiency. Our experimental results suggest that the method is stable under its hyper-parameters and that the ranking objective by Eq.~\ref{eq:objective} helps finding efficient architectures. One limitation is the high variability in the selected architectural components and their parameter counts. Future work should investigate how this variability can be reduced, e.g., by combining PainNAS with meta-learning~\cite{hihn2020specialization} to select architectures that rapidly adapt to unseen subjects, followed by lightweight continual learning~\cite{hihn2023hierarchically,hihn2024online} to handle subject-specific physiological changes over time. Another promising research avenue is the design of dedicated NAS algorithms for LOSO settings to capture architectural and modality interdependencies.

\section*{\uppercase{Acknowledgements}}
Generative AI was used to assist in writing of this manuscript (OpenAI ChatGPT) and in developing the code for the experiments (OpenAI Codex). All parts created by an AI have been critically examined and reviewed by the authors.
\bibliographystyle{apalike}
{\small
\bibliography{main}}

\end{document}